\documentclass[10pt,twocolumn,letterpaper]{article}

\usepackage{iccv}              %

\usepackage{tabularx}
\newcolumntype{Y}{>{\centering\arraybackslash}X}

\colorlet{colorFst}{Green!25}       %
\colorlet{colorSnd}{SpringGreen!45} %
\colorlet{colorTrd}{Yellow!30}      %
\colorlet{colorLow}{darkgray!30}    %

\definecolor{iccvblue}{rgb}{0.21,0.49,0.74}
\usepackage[pagebackref,breaklinks,colorlinks,allcolors=iccvblue]{hyperref}

\usepackage{amsmath}
\usepackage{adjustbox}
\usepackage{booktabs}
\usepackage{pifont}
\usepackage{xcolor}
\usepackage[table]{xcolor}
\usepackage{multirow}
\usepackage{graphicx}
\usepackage{subcaption}
\usepackage[accsupp]{axessibility}  %

\newcommand{\ourmodel}{SatDiFuser}

\def\confName{ICCV}
\def\confYear{2025}

\title{Can Generative Geospatial Diffusion Models Excel as Discriminative Geospatial Foundation Models? }

\author{%
Yuru Jia\textsuperscript{1,2} \quad
Valerio Marsocci\textsuperscript{3} \quad
Ziyang Gong\textsuperscript{4}  \quad
Xue Yang\textsuperscript{5} \\
Maarten Vergauwen\textsuperscript{1} \quad
Andrea Nascetti\textsuperscript{2}
\vspace{5px}
\\
{\small \textsuperscript{1}KU Leuven} \quad
{\small \textsuperscript{2}KTH} \quad
{\small \textsuperscript{3}European Space Agency} \quad
{\small \textsuperscript{4}Shanghai AI Lab} \quad
{\small \textsuperscript{5}SJTU} \\
\vspace{-20px}
}

\begin{document}
\maketitle
\begin{abstract}

Self-supervised learning (SSL) has revolutionized representation learning in Remote Sensing (RS), advancing Geospatial Foundation Models (GFMs) to leverage vast unlabeled satellite imagery for diverse downstream tasks. 
Currently, GFMs primarily employ objectives like contrastive learning or masked image modeling, owing to their proven success in learning transferable representations. However, generative diffusion models, which demonstrate the potential to capture multi-grained semantics essential for RS tasks during image generation, remain underexplored for discriminative applications. This prompts the question: can generative diffusion models also excel and serve as GFMs with sufficient discriminative power? In this work, we answer this question with \textbf{\ourmodel}, a framework that transforms a diffusion-based generative geospatial foundation model into a powerful pretraining tool for discriminative RS.
By systematically analyzing multi-stage, noise-dependent diffusion features, we develop three fusion strategies to effectively leverage these diverse representations.
Extensive experiments on remote sensing benchmarks show that {\ourmodel} outperforms state-of-the-art GFMs, achieving gains of up to +5.7\% mIoU in semantic segmentation and +7.9\% F1-score in classification, demonstrating the capacity of diffusion-based generative foundation models to rival or exceed discriminative GFMs. 
The source code is available at: \href{https://github.com/yurujaja/SatDiFuser}{https://github.com/yurujaja/SatDiFuser}.
\end{abstract}
\

\section{Introduction}
\label{sec:intro}

\begin{figure}[t]
    \centering
    \includegraphics[width=\linewidth]{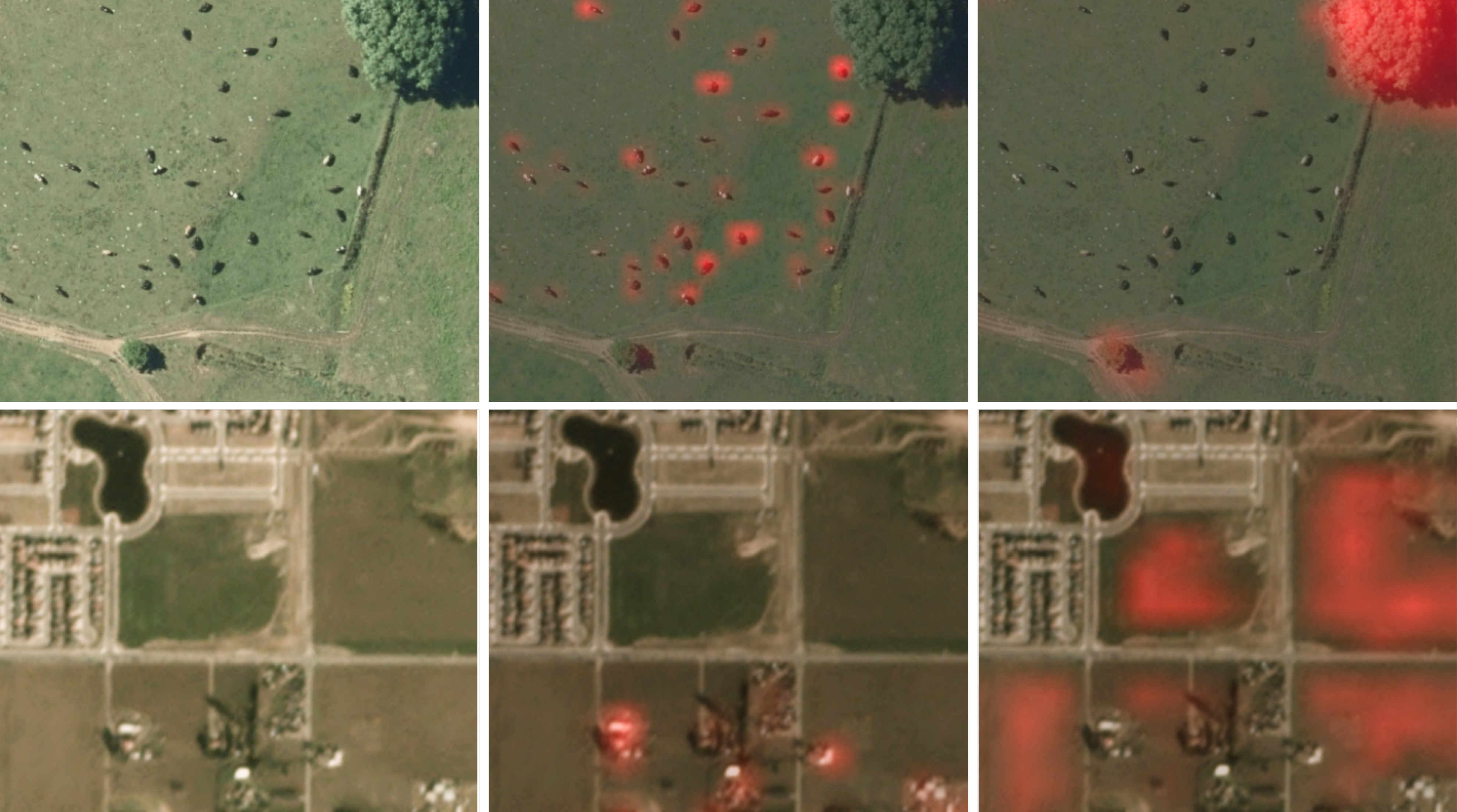}
    \caption{\textbf{Self-attention maps} from an \textbf{off-the-shelf} geospatial generative diffusion model~\cite{khanna2024diffusionsat} on satellite images. Semantically similar objects strongly attend to each other at different scales, highlighting the potential of generative diffusion models for discriminative tasks in remote sensing.}
    
    \label{fig:cattle_sa}
    \vspace{-7pt}
\end{figure}

Self-supervised learning (SSL) has emerged as a pivotal paradigm in computer vision, enabling models to learn robust representations without relying on labeled data. This capability is especially valuable for remote sensing (RS), where vast amounts of unlabeled satellite imagery can be leveraged for downstream tasks like land-cover classification and change detection~\cite{AI4EO-tuia,Mai2022TowardsAF}. Modern SSL frameworks, such as contrastive learning~\cite{chen2020simple,radford2021learning}, self-distillation~\cite{grill2020bootstrap,caron2021emerging}, and masked image modeling (MIM)~\cite{he2022masked,xie2022simmim}, have driven substantial progress in developing Geospatial Foundation Models (GFMs)~\cite{wang2022self,marsocci2024pangaea}, significantly advancing RS image analysis.
Despite this progress, prevalent SSL paradigms exhibit inherent limitations under the RS context. Contrastive learning relies on constructing informative positive and negative pairs, a challenging task in complex RS imagery, and its global instance-level supervision tends to overlook spatially fine-grained details. Similarly, MIM’s patch-level reconstruction objective may produce overly coarse representations, especially in many RS scenes dominated by homogeneous regions, where masking becomes trivial and limits learning effectiveness. These drawbacks restrict the adaptability of current SSL approaches to more complex and multi-scale geospatial data, as recently revealed by~\cite{marsocci2024pangaea}. Motivated by these limitations, we explore an alternative yet underexplored SSL framework in geospatial domains: generative diffusion models. Diffusion models inherently perform self-supervision through a denoising process that models the underlying data distribution, offering a promising pathway toward capturing richer semantic representations from satellite imagery.

Diffusion models~\cite{ho2020denoising,song2021denoising,Rombach_2022_CVPR} have demonstrated extraordinary image generation capabilities by corrupting data with noise in a forward process and learning a reverse process to recover original data. 
Recent efforts~\cite{khanna2024diffusionsat,metaearth,tang2024cross,liu2025text2earth} have explored diffusion-based generative foundation models to produce high-fidelity RS scenes. While diffusion models have primarily been adopted for image synthesis, emerging works prove that these generative approaches can learn meaningful semantic representations~\cite{fuest2024diffusion,xu2023open,luo2023dhf,couairondiffcut}.
We further hypothesize that diffusion models provide distinct advantages for representation learning in RS imagery. During iterative denoising, these models simultaneously consider both global semantic structures and fine-grained local details, —necessary for coherent image synthesis. 
This aligns particularly well with RS data, where images typically contain objects and regions spanning various scales.
As illustrated by~\cref{fig:cattle_sa}, pretrained diffusion models naturally form meaningful self-attention patterns at multiple scales, from sparse cattle pixels to broader objects like trees and agricultural fields. Additionally, diffusion models explicitly model noisy data distributions, potentially providing robustness against sensor noise and atmospheric interference commonly present in RS data, enabling the extraction of more reliable representations~\cite{liu2024diffusion}.

These advantages indicate that generative diffusion models have untapped potential for discriminative tasks. However, one of the challenges preventing the widespread adoption of diffusion models for such tasks, both in RS and CV, is the lack of a unified and effective framework to leverage these fruitful features. Existing approaches adapt diffusion-derived representations differently depending on the task, leading to inconsistent performance and limiting broader applicability. In this work, we bridge this gap by repurposing diffusion-based generative models for self-supervised representation learning in RS. We introduce \textbf{{\ourmodel}}, a flexible framework designed to efficiently harness multi-stage diffusion features, unlocking their full discriminative power for various RS tasks.

Specifically, we systematically analyze how noise-level–dependent features across diffusion stages affect performance on diverse RS tasks. 
To mitigate task-dependent feature selection, 
{\ourmodel} hierarchically explore three feature fusion strategies: (i) a global weighted fusion for a broad aggregation, (ii) a localized weighted approach for input-dependent, fine-grained selection, and (iii) a mixture-of-experts design jointly modeling inter-timestep and inter-module relationships. When benchmarked against top GFMs pretrained via alternative objectives, {\ourmodel} demonstrates superior accuracy on classification and semantic segmentation tasks, confirming the efficacy of generative diffusion models as a powerful SSL framework for RS. While we employ DiffusionSat~\cite{khanna2024diffusionsat}—a latent diffusion model (LDM) pretrained at scale on satellite imagery—as our backbone, {\ourmodel} can be extended to other diffusion architectures, laying the foundation for a broader integration of diffusion-based generative modeling into geospatial analysis.

In summary, our \textbf{contributions} are threefold: \textbf{First}, to the best of our knowledge, we are the first to comprehensively adapt a large-scale diffusion-based generative model for self-supervised representation learning in RS, forming a diffusion-driven GFM. 
\textbf{Second}, we propose three efficient multi-stage feature fusion strategies, offering global weighted fusion, localized weighted fusion, and a mixture-of-experts fusion to maximize the discriminative power of diffusion-based features. 
\textbf{Third}, by benchmarking {\ourmodel} against leading GFMs across various RS tasks, we show that diffusion-driven GFMs offer notable advantages, paving the way for broader synergies between diffusion-based generative modeling and discriminative geospatial analysis.

\section{Related Work}
\label{sec:related}

\paragraph{Diffusion Models for Representation Learning.}
A number of recent works~\cite{baranchuk2022labelefficient,xu2023open,xiang2023denoising,wu2023datasetdm,mukhopadhyay2025text,luo2023dhf,tian2024diffuse,couairondiffcut,khani2023slime,zhang2025three,fuest2024diffusion} in computer vision have started investigating the discriminative representations inherent in pretrained diffusion models. 
Some methods mine self-attention tensors for unsupervised segmentation, either by merging attention maps (DiffSeg~\cite{tian2024diffuse}) or by constructing affinity graphs (DiffCut~\cite{couairondiffcut}). Others incorporate cross-attention signals (SLiMe~\cite{khani2023slime}), fine-tuning text embeddings to segment objects at varied granularity. 
Meanwhile, DatasetDM~\cite{wu2023datasetdm} extracts multi-scale features from a Stable Diffusion UNet to train a dataset-generation model capable of producing densely annotated images.
Diffusion HyperFeatures~\cite{luo2023dhf} further enhances feature aggregation by incorporating multi-timestep feature maps, creating a feature descriptor for semantic keypoint correspondence tasks. Additionally, 
REPA~\cite{yu2024repa} demonstrates the improving synergies between representation learning and generative models by utilizing external high-quality representations. Inspired by these efforts, we exploit diffusion models that are trained on large-scale global satellite imagery, adapting their representational capacity to a wide range of RS tasks.

\paragraph{SSL for Remote Sensing.}
Supervised pretrained RS models (e.g.,~\cite{bastani2023satlaspretrain}) require extensive labeled data, which can be costly to obtain at scale. To circumvent this limitation, SSL has greatly advanced deep learning in RS by leveraging abundant unlabeled satellite imagery. Early efforts, such as SSL4EO-L~\cite{stewart2024ssl4eo} and SSL4EO-S12~\cite{ssl4eo}, introduced globally distributed Landsat-8 and Sentinel-1/2 data, which have been used to train state-of-the-art SSL models like MAE~\cite{he2022masked} and DINO~\cite{caron2021emerging}.
To address the unique characteristics of RS data, numerous studies have integrated RS-specific features, such as spatiotemporal embeddings and multi-spectral information, into SSL frameworks. These include masked image modeling (MIM)-based approaches (e.g., SatMAE~\cite{cong2022satmae}, Scale-MAE~\cite{reed2023scalemae}, DOFA~\cite{xiong2024neural}), contrastive frameworks (e.g., GASSL~\cite{ayush2021geography}, CROMA~\cite{fuller2023croma}, SkySense~\cite{guo2024skysense}), and self-distillation methods (such as~\cite{tolan2024very,marsocci2024crosssensor}). Additionally, various multi-modal methods~\cite{astruc2024omnisat,Hong2024spectralgpt,han2024bridging} extend these techniques by incorporating diverse RS modalities. Other learning strategies, including continual pretraining~\cite{mendieta2023gfm} and multi-task pretraining~\cite{wang2024mtp,gong2024crossearth}, have also been explored to better adapt to satellite data.
Despite this variety of RS-focused SSL methods, diffusion models remain largely unexplored as an SSL pretraining strategy. This work seeks to explore this promising direction.

\paragraph{Diffusion Models in RS.}
Diffusion models have gained increased traction in RS, being applied to image generation, enhancement, and interpretation~\cite{liu2024diffusion}. 
Recent efforts~\cite{khanna2024diffusionsat,metaearth,zheng2024changen2,tang2024crs,toker2024satsynth,espinosa2025cop} have focused on developing diffusion-based generative foundation models for high-fidelity satellite image synthesis. For instance, DiffusionSat~\cite{khanna2024diffusionsat} generates data conditioned on semantic text and metadata, while MetaEarth~\cite{metaearth} enables arbitrary-sized image generation using a resolution-guided approach. 
Beyond synthesis, numerous diffusion-based methods address image enhancement tasks, including denoising~\cite{he2023tdiffde,pang2024hir}, cloud removal~\cite{wang2024idf,zou2024diffcr}, and super-resolution~\cite{wang2025semantic,dong2024building}, showcasing their versatility in RS.
Another line of research focuses on discriminative applications \cite{le2024detecting}, though these often rely on labeled data and are limited to specific tasks, such as semantic segmentation~\cite{amit2021segdiff, kolbeinsson2024multi, zhou2024exploring, li2024mdfl, qu2024lds2ae} or change detection~\cite{wen2024gcd, tian2024swimdiff, zhang2023diffucd, jia2024siamese}. For example, SegDiff~\cite{amit2021segdiff} diffuse ground-truth masks, while others use class predictions or labeled guidance~\cite{kolbeinsson2024multi,li2024mdfl,qu2024lds2ae}. Although a few studies have explored diffusion as a label-free pretraining framework, they remain narrowly focused on a single application scenario, such as hyperspectral images segmentation~\cite{zhou2024exploring,sigger2024unveiling}, or change detection~\cite{bandara2022ddpm}. In contrast, our work provides a comprehensive investigation of the discriminative capabilities of diffusion-based generative models across multiple RS tasks. By moving beyond task-specific solutions and limited testing, our method advances the broader potential of diffusion-driven GFMs pretrained on global-scale data.

\section{Methods}
\label{sec:method}
\begin{figure*}[tb!]
    \centering
    \includegraphics[width=\linewidth]{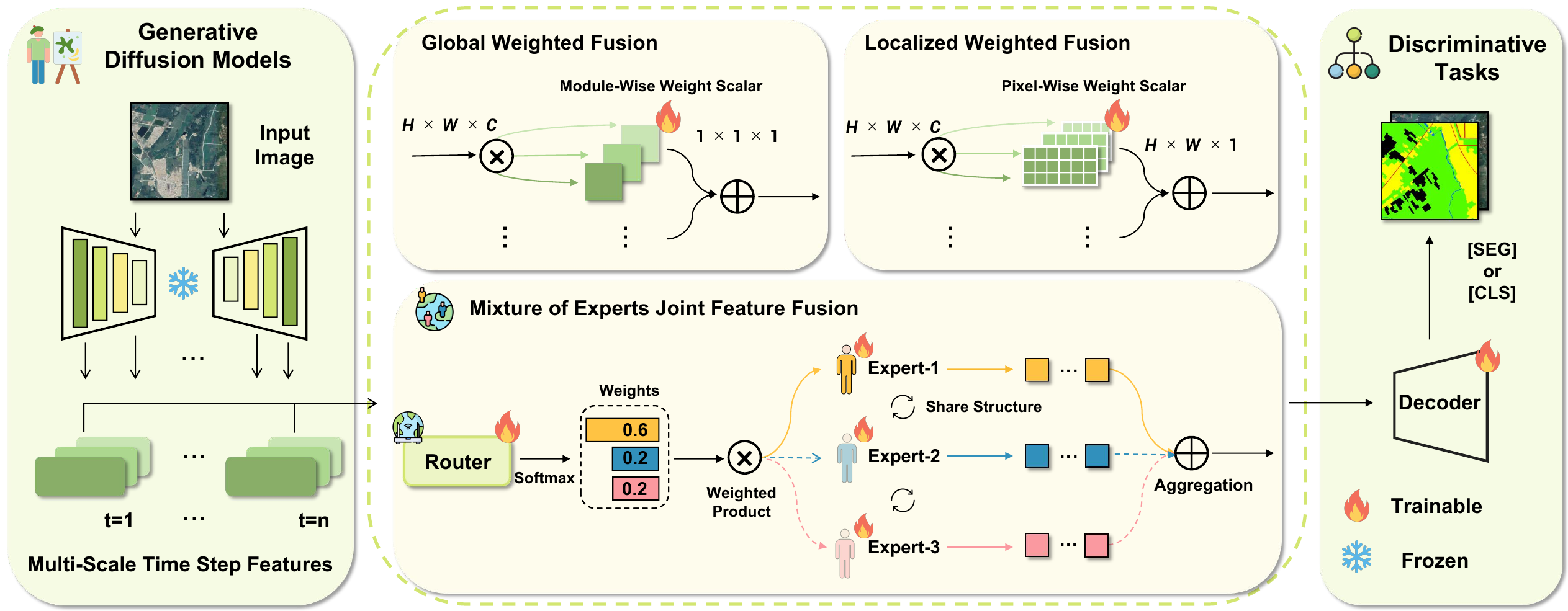}
    \caption{\textbf{Method Overview.} Our \textbf{\ourmodel} framework leverages diffusion-based generative foundation models as self-supervised feature extractors for downstream discriminative remote sensing tasks. \textbf{(Left)}: The pre-trained geospatial diffusion model captures diverse representations at different scales and timesteps for satellite imagery. \textbf{(Right)}: Three feature fusion strategies are explored to effectively leverage these features: (1) Global Weighted Fusion applies learnable module-wise scalars for broad aggregation. (2) Localized Weighted Fusion learns pixel-wise weights for spatially varying importance. (3) Mixture of Experts (MoE) Joint Fusion uses specialized experts to model complex feature interactions. The fused features are fed into a task-specific decoder for different RS tasks.}
    \label{fig:method}
    \vspace{-10pt}
\end{figure*}

Our approach builds on DiffusionSat~\cite{khanna2024diffusionsat} - a satellite-adapted LDM based on Stable Diffusion v2-1~\cite{Rombach_2022_CVPR}\footnote{Currently, DiffusionSat is the only openly available large-scale generative geospatial foundation model with pretrained weights and accessible training/inference code.}. We first conduct an overview of the key internal components of the diffusion model and demonstrate the extraction of multi-scale multi-timesteps features in \cref{sec:feat-extract}. We then propose three fusion strategies to systematically aggregate these features, i.e., via global weighted fusion (\cref{sec:global-weight}), via localized weighted fusion(\cref{sec: adaptive-weights}), and via a mixture-of-experts mechanism (\cref{sec: moe}). An overview of our method is illustrated in~\cref{fig:method}.

\subsection{Feature Extraction from Diffusion Process}
\label{sec:feat-extract}
In an LDM~\cite{Rombach_2022_CVPR}, an input image $\mathbf{x}$ is first mapped into a latent representation  $\mathbf{z}\in\mathbb{R}^{H_0\times W_0\times C_0}$ via an autoencoder. 
To extract features from the diffusion process, we start with the clean latents $\mathbf{z}$ and employ DDIM inversion~\cite{song2021denoising} to trace a reverse noise path, obtaining noisy latents. We then run the denoising diffusion model on these noisy latents to capture multi-scale multi-timestep feature maps. This inversion approach yields faithful latent representations, helping preserve fine-grained details in deterministic tasks. Complete equations are provided in the Supplementary~\cref{supp:diff-equations}.

\paragraph{Backbone Architecture.} 
\label{sec: model-arch}
The denoising backbone follows a U-Net-like architecture that generates features at $S=4$ scales with resolutions
$\left\{\frac{H_0}{2^{s-1}} \times \frac{W_0}{2^{s-1}}\right\}_{s=1}^S$. Each scale contains multiple \textit{residual blocks} capturing local spatial information, and \textit{transformer blocks} including a self-attention (SA) and a cross-attention (CA) mechanism. The SA block captures contextual dependencies within the latent itself, while the CA block encodes interactions between the latent and additional conditioning signals (e.g., text prompts).
Across the diffusion process, each noise level is conditioned on a \textit{timestep} $t \in \{1,\dots,T\}$. At each $t$, the U-Net refines the noisy latent toward a cleaner state. 
This procedure naturally produces a variety of spatiotemporal features.

For simplicity, we denote the \textit{SA} outputs at scale $s$ and timestep $t$ by $\mathbf{A}_{t, \mathrm{s}}\in\mathbb{R}^{h_s\times w_s\times d_s^a}$, the \textit{CA} outputs by $\mathbf{C}_{t, \mathrm{s}}\in\mathbb{R}^{h_s\times w_s\times d_s^c}$, and the \textit{ResNet} residual outputs by $\mathbf{R}_{t, \mathrm{s}}\in\mathbb{R}^{h_s\times w_s\times d_s^r}$, 
where $h_s=H_0/2^{s-1},w_s=W_0/2^{s-1}$, and $d_s^a, d_s^c, d_s^r$ are channel dimensions that may vary across blocks. Note that for attention blocks, we recover spatial dimensions for outputs to maintain consistency with the ResNet outputs.

These multi-scale, multi-timestep features form the building blocks for subsequent recognition tasks, as they embed both coarse- and fine-grained cues from different stages of the diffusion process. A straightforward approach to utilize these features is to attach a task‐specific decoder on top of any desired subset. However, effectively navigating which blocks and timesteps to pick can be cumbersome, and a simple concatenation often yields marginal improvements (see \cref{sec:ablations}). 
To address this, we propose three feature fusion strategies in the following sections that combine diverse features effectively to optimize downstream task performance.

\subsection{Global Weighted Feature Fusion}
\label{sec:global-weight}

Inspired by Diffusion Hyperfeatures~\cite{luo2023dhf}, we adopt a learnable global‐weight aggregation scheme across timesteps and feature blocks. However, unlike~\cite{luo2023dhf}, which resizes all features to a uniform resolution, we maintain the original multi‐scale resolutions, resulting in a feature pyramid $\{\mathbf{X}_1, \mathbf{X}_2, \dots, \mathbf{X}_S\}$, allowing the features to capture the multi-scale nature inherent in RS images. At each scale $s$, the final aggregated feature is a weighted sum:
\begin{equation}
\mathbf{X}_s
\;=\;
\sum_{t=1}^{T_{sel}} \sum_{l=1}^{L_{sel}} w_{l, t} \cdot \Phi_s^l(\mathbf{F}_{t,s}^l),
\end{equation}
where $\mathbf{F}_{t,s}^l$ denotes the $l$-th feature block at scale $s$ and timestep $t$, and $w_{l,t}$ is a learnable weight scalar for each block‐timestep pair. $\mathbf{A}$, $\mathbf{C}$, $\mathbf{R}$ are examples of possible feature types, and $\Phi$ is a projection network that aligns channels. $T_{sel}$ and $L_{sel}$ indicate the total number of selected timesteps and feature blocks. 

By learning global importance weights, this method offers a simple and efficient way to integrate multi-scale, multi-timestep diffusion features, capturing a broad representation while minimizing additional computational costs.

\subsection{Localized Weighted Feature Fusion}
\label{sec: adaptive-weights}

Unlike global weighting, which applies a uniform mixing factor to entire feature maps, we then investigate \emph{pixel‐level} weighting that dynamically emphasizes different features at each spatial location.

Specifically, for a given scale $s$, we first compute a \emph{reference feature} by averaging the extracted feature maps $\{\mathbf{F}_{t,s}^l\}$. This reference is fed into a lightweight gating function (e.g., a small convolutional network) to generate pixel‐wise weights $\{\mathbf{W}_{t,s}^l\}\in \mathbb{R}^{h_s\times w_s}$. 
These weights are then normalized and applied to the corresponding features:
\begin{equation}
\mathbf{X}_s(u,v)
\;=\;
\sum_{t=1}^{T_{sel}} \sum_{l=1}^{L_{sel}}
\mathbf{W}_{t,s}^l(u,v)  \cdot \Phi_s^l(\mathbf{F}_{t,s}^l)(u,v),
\end{equation}
where $(u,v)$ denotes a spatial location. Repeating the same gating process at each scale produces a pyramid of pixel‐wise fused features.

By allowing a more nuanced feature aggregation, this spatially adaptive scheme can preserve local details more effectively. Its fine‐grained emphasis is particularly suited to objects with intricate outlines or heterogeneous textures in RS images,
offering richer spatial detail than a single global weighting factor. However, this sensitivity can also respond strongly to local variations such as illumination differences (see~\cref{sec:analysis}).

\subsection{MoE Joint Feature Fusion}
\label{sec: moe}
The previous fusion methods explicitly decouple each feature map, encouraging the model to learn patterns for “which modules within which timestep to emphasize”. In contrast, to model the more complex interactions between timesteps and feature blocks, we introduce a joint modeling method using a mixture-of-experts mechanism. 

Mixture of Experts (MoE) is a sparsely activated architecture that partitions the model’s parameters into expert sub-networks, coordinated by a routing function that selects which experts to activate~\cite{shazeer2017outrageously}. This divide-and-conquer approach allows the model to tackle complex tasks by assigning specialized experts to different data aspects. This capability is especially advantageous in remote sensing, where images display diverse patterns, ranging from fine-grained textures to large-scale contextual variations.

Building on this idea, we adapt the MoE paradigm to fuse diffusion features at each scale by jointly modeling different module outputs and multiple timesteps. Specifically, for each selected timestep $t\in\{1,\dots,T_{sel}\}$ at scale $s$, we first concatenate the module‐specific features $\{\mathbf{F}_{t,s}^l\}$ into a single vector $\mathbf{X}_{t,s}$ along the channel dimension:
\begin{equation}
\mathbf{X}_{t,s} = \mathrm{Concat}\bigl(\mathbf{F}_{t,s}^1,\dots,\mathbf{F}_{t,s}^{L_{sel}}\bigr)
\in \mathbb{R}^{B\times C_s \times H_s \times W_s},
\end{equation}
where $C_s$ is the total channel dimension after concatenation. A shared MoE layer $f_{\mathrm{MoE}}(\cdot)$ then processes $\mathbf{X}_{t,s}$ via $E$ expert sub‐networks $\{f_1,\dots,f_E\}$ and a gating function $\gamma$. Formally,
\vspace{-10pt}
\begin{equation}
\mathbf{Y}_{t,s}
\;=\;
f_{\mathrm{MoE}}\bigl(\mathbf{X}_{t,s}\bigr)
\;=\;
\sum_{e=1}^{E} \gamma_e\!\bigl(\mathbf{X}_{t,s}\bigr)\;f_e\!\bigl(\mathbf{X}_{t,s}\bigr).
\end{equation}
Each expert $f_e$ focuses on certain patterns in the concatenated features, while $\gamma_e(\mathbf{X}_{t,s})$ indicates how strongly to activate the experts. Optionally, top‐$k$ routing \cite{shazeer2017outrageously} can reduce computational overhead by zeroing out less relevant experts. After processing each timestep, we sum the resulting outputs across all selected $t$ to obtain $\mathbf{X}_s$.

Compared to scalar or pixel‐wise weighting, this joint formulation explicitly captures the synergy among different network modules and timesteps. By leveraging specialized sub-networks to capture diverse diffusion features, it offers a robust and flexible representation that can adapt to varied patterns in RS data.

\section{Experiments}
\label{sec:experiments}

\subsection{Evaluation Protocol}
\label{sec:experiment-protocal}
To assess the discriminative power of generative diffusion features and validate the effectiveness of feature fusion strategies of {\ourmodel}, we perform evaluations on a diverse set of classification and semantic segmentation tasks. Following standard evaluation protocols in recent GFMs~\cite{marsocci2024pangaea,xiong2024neural}, we freeze the pretrained generative backbone, and only train {\ourmodel} components with task-specific decoders: a linear head for classification tasks and a UPerNet decoder~\cite{xiao2018unified} for segmentation tasks.

We uniformly employ this setting to benchmark against state-of-the-art large-scale pretrained RS models, covering diverse pretraining paradigms, including MIM~\cite{reed2023scalemae,xiong2024neural,ssl4eo,mendieta2023gfm}, contrastive learning~\cite{mall2024remote,fuller2023croma}, self-distillation~\cite{ssl4eo}, and supervised pretraining~\cite{bastani2023satlaspretrain}. Detailed feature extraction settings for these models are provided in~\cref{sec:implementation-detail}. We also include a fully supervised ConvNeXt~\cite{liu2022convnet} for classification and a UNet~\cite{ronneberger2015u} for segmentation for reference. To handle discrepancies between the spectral bands available in the datasets and those required during GFMs pretraining, we follow the standard practice in~\cite{marsocci2024pangaea} by matching available bands and zero‐filling any missing ones for all the models. Specifically, we only match RGB bands for DiffusionSat across all tasks.

\vspace{-3pt}
\paragraph{Downstream Tasks.} We adopt GEO-Bench~\cite{lacoste2024geo}, which contains semantic segmentation and classification datasets, covering diverse application domains (e.g., agriculture, urban, forest, etc.) and geographic regions. Dataset-specific details, including dataset sizes and spectral properties, are summarized in the supplementary material (\cref{supp:datsets-details}).

\vspace{-3pt}
\paragraph{Pretrained DiffusionSat.} The original DiffusionSat model~\cite{khanna2024diffusionsat} supports text and metadata conditioning during image synthesis. In our experiments, we omit metadata and class-specific conditioning to avoid potential information leakage.  For text prompts, we use a generic phrase, \emph{``A satellite image’’}, to keep the conditioning consistent across all tasks.

\subsection{Implementation Details}
\label{sec:implementation-detail}

For DiffusionSat, we select ResNet and self-attention outputs at timesteps $\{1, 100, 200\}$ from the decoder blocks of its UNet (details justified in~\cref{sec:ablations}). For comparison GFMs, we extract features from evenly spaced layers based on the specific GFM architecture, following common protocols~\cite{marsocci2024pangaea, reed2023scalemae}. For instance, in a 12-layer ViT-based MAE model, we select features from layers indexed at (3, 5, 7, 11).
All models are optimized using AdamW with an initial learning rate of 0.01, scheduled with cosine decay after a 5-epoch warm-up. Images are cropped or resized to match the pretraining resolution required by each GFM; specifically for DiffusionSat, which offers models trained on 512px and 256px resolutions, we resize images to these dimensions using bilinear interpolation based on their original resolution. Each spectral band is normalized individually using the minimum and maximum values calculated across the entire dataset. We use a training batch size of 32. Additional dataset-specific implementation details, such as loss functions and training epochs, are provided in the supplementary material (\cref{supp:eval-details}).

\subsection{Main Results}
\label{sec:main-results}
\begin{table}[t!]
    \centering\renewcommand{\arraystretch}{1.05}
    \setlength{\tabcolsep}{3pt}
    \resizebox{1.0\columnwidth}{!}{
    \begin{tabular}{lcccccc}
        \toprule
             Method & pv-s & nz-c & neon & cashew & sa-c & ches \\
        \midrule

        \rowcolor[gray]{0.9}
        Fully Supervised &  94.7	&85.1	&64.2	&79.9	&34.4	&70.4 \\
        Satlas~\cite{bastani2023satlaspretrain}&92.3	&83.1	&52.0	&49.1	&31.6	&52.2 \\
        SSL4EO-MAE~\cite{ssl4eo} &89.2	&78.7	&53.1	&57.8	&28.6	&52.0\\
        ScaleMAE~\cite{reed2023scalemae} & 94.2	&\underline{\colorbox{colorFst}{\bf 84.1}}	&55.9	&47.8	&20.1	&61.1\\
        SSL4EO-DINO~\cite{ssl4eo} &89.0	&78.9	&53.7	&61.3	&31.6	&54.9 \\ 
        GFM~\cite{mendieta2023towards} &93.1	&82.4	&54.5	&53.5	&25.0	&63.7\\ 
        RemoteCLIP~\cite{mall2024remote} &93.2	&80.7	&55.5	&51.7	&22.1	&55.2\\
        CROMA~\cite{fuller2023croma}&92.5	&83.4	&56.3	&\underline{62.2}	&\underline{\colorbox{colorSnd}{32.3}}	&63.6 \\
        DOFA~\cite{xiong2024neural} &\underline{94.8}	&82.8	&\underline{58.1}	&53.9	&26.6	&\underline{65.7}\\
        
        \midrule
        \multicolumn{7}{l}{\cellcolor[HTML]{EEEEEE}{\textit{\ourmodel (Ours)}}} \\
        Global fusion  &\colorbox{colorSnd}{95.1}	&\colorbox{colorTrd}{83.5}	&\colorbox{colorTrd}{61.8}	&\colorbox{colorFst}{\bf 66.5}	&\colorbox{colorFst}{\bf 32.6}	&\colorbox{colorTrd}{69.5}\\
        Localized  fusion & \colorbox{colorTrd}{95.0}	&83.2	&\colorbox{colorFst}{\bf 63.8}	&\colorbox{colorTrd}{64.8}	&31.9	&\colorbox{colorSnd}{ 70.3}\\
        MoE fusion  & \colorbox{colorFst}{\bf 95.3}	& \colorbox{colorSnd}{83.7}	& \colorbox{colorSnd}{63.4}	& \colorbox{colorSnd}{ 66.1}	& \colorbox{colorTrd}{31.9}	& \colorbox{colorFst}{\bf 71.6}
        \\

          & \textcolor{ForestGreen}{$\uparrow$ \textbf{{0.5}}}	& 	$\downarrow$ \textbf{{0.4}}&\textcolor{ForestGreen}{$\uparrow$ \textbf{{5.7}}}& \textcolor{ForestGreen}{$\uparrow$ \textbf{{4.3}}}	& \textcolor{ForestGreen}{$\uparrow$ \textbf{{0.3}}} & \textcolor{ForestGreen}{$\uparrow$ \textbf{{5.9}}} \\

        \bottomrule
    \end{tabular}
    }
    \caption{\textbf{Semantic segmentation performance} with the UPerNet decoder, reported as mIoU $\uparrow$ (in \%). The top three results are highlighted as \colorbox{colorFst}{\bf first}, \colorbox{colorSnd}{second}, and \colorbox{colorTrd}{third}. The best results among other pretrained remote sensing models are \underline{underlined}. The final row presents the performance difference between our best results and the top-performing pretrained RS models. Abbreviations: pv-s, nz-c, neon, cashew, sa-c, and ches correspond to m-pv4ger-seg, m-nz-cattle, m-NeonTree, m-cashew-plantation, m-SA-crop-type, and m-chesapeake-landcover, respectively.}
    \label{tab:ss_comparison}
    \vspace{-3pt}
\end{table}

In~\cref{tab:ss_comparison}, we evaluate the semantic segmentation performance of {\ourmodel} using three distinct fusion strategies, comparing it against other pretrained RS models. {\ourmodel} achieves the highest mIoU on five of the six tasks, with particularly large gains on m-NeonTree, m-cashew-plantation, and m-chesapeake-landcover (5.7\%, 4.3\%, and 5.9\% improvements over the strongest baselines, respectively). Notably, even on multi-spectral datasets like m-cashew-plan and m-SA-crop-type, {\ourmodel} maintains superior performance using only RGB bands, underscoring its robustness. These results highlight the strong discriminative knowledge embedded in large-scale pretrained generative diffusion models and validate the effectiveness of our approach in transferring that knowledge to downstream dense prediction tasks.

\begin{table}[t!]
    \centering\renewcommand{\arraystretch}{1.1}
    \setlength{\tabcolsep}{4pt}
    \resizebox{1.0\columnwidth}{!}{
    \begin{tabular}{lcccccc}
        \toprule
        Method & ben & bk & es & fn  & pv & s2s  \\
        \midrule

        \rowcolor[gray]{0.9}
        Fully Supervised &  69.4 & 98.9	&97.7	&56.8	&98.0	&58.1 \\
        Satlas~\cite{bastani2023satlaspretrain} & 54.3 &88.7	&92.1	&40.8	&95.2	&55.3 \\
        SSL4EO-MAE~\cite{ssl4eo}  & 46.0 & 92.0 & 82.4 & 40.5 & 90.9 & 47.3 \\
        ScaleMAE~\cite{reed2023scalemae} & 39.8 & 90.3	&78.9	&45.6	&96.9	&21.8 \\
        SSL4EO-DINO~\cite{ssl4eo}   & 46.4 & 91.8 & 80.9 & 40.2 & 91.3 & 44.7 \\
        GFM~\cite{mendieta2023towards} & 48.9 &95.7	&89.2	&52.4	&96.8	&46.3 \\
        RemoteCLIP~\cite{mall2024remote}& 43.4 &95.5	&85.5	&51.8	&96.0 & 38.8\\
        CROMA~\cite{fuller2023croma}  & \underline{58.3} &\underline{95.8}	&92.6	&49.5	&93.6	&53.8\\        
        DOFA~\cite{xiong2024neural} &  50.8 & 95.3	&\underline{93.8}	&\underline{52.5}	& \underline{97.3}	&\underline{55.4} \\

        \midrule
        \multicolumn{7}{l}{\cellcolor[HTML]{EEEEEE}{\textit{\ourmodel (Ours)}}} \\
        Global fusion  & \colorbox{colorFst}{\bf 66.2} & \colorbox{colorSnd}{98.1} & \colorbox{colorFst}{\bf 97.7} & \colorbox{colorSnd}{57.1} & \colorbox{colorTrd}{97.4} & \colorbox{colorFst}{\bf 59.3} \\
        Localized  fusion & \colorbox{colorTrd}{64.8} & \colorbox{colorTrd}{97.7} & \colorbox{colorTrd}{96.8} & \colorbox{colorTrd}{56.2} & \colorbox{colorSnd}{97.7} & \colorbox{colorSnd}{58.9} \\
        MoE fusion  & \colorbox{colorSnd}{65.8}  & \colorbox{colorFst}{\bf 98.3} & \colorbox{colorSnd}{97.3} & \colorbox{colorFst}{\bf 58.4} & \colorbox{colorFst}{\bf 97.8} &  \colorbox{colorTrd}{58.8} \\

          & \textcolor{ForestGreen}{$\uparrow$ \textbf{{7.9}}}	& \textcolor{ForestGreen}{$\uparrow$ \textbf{{2.5}}}	&\textcolor{ForestGreen}{$\uparrow$ \textbf{{3.9}}}& \textcolor{ForestGreen}{$\uparrow$ \textbf{{5.9}}}	& \textcolor{ForestGreen}{$\uparrow$ \textbf{{0.5}}} & \textcolor{ForestGreen}{$\uparrow$ \textbf{{3.9}}} \\
        \bottomrule
    \end{tabular}
    }
    \caption{\textbf{Classification performance} using a linear classifier. The reported metric is the F1 score for the m-bigearthnet dataset and top-1 accuracy (\%) $\uparrow$ for all other datasets. Our method consistently achieves the highest performance. Abbreviations: ben, bk, es, fn, pv and s2s correspond to m-bigearthnet, m-brick-kiln, m-eurosat, m-forestnet, m-pv4ger, and m-so2sat, respectively.}
    \label{tab:cls_comparison}
\end{table}

Turning to classification, \cref{tab:cls_comparison} shows that {\ourmodel} consistently outperforms competing RS models across the benchmark.  The gains are especially pronounced on challenging datasets that are not saturated, with improvements of up to +7.9\% on m-bigearthnet, +5.9\% on m-forestnet, and +3.9\% on m-so2sat. Especially, {\ourmodel} even surpasses fully supervised settings on m-forestnet and m-so2sat datasets. These results underscore the versatility of diffusion-based features in RS tasks and confirm the potential of {\ourmodel} as a robust foundation for both dense prediction and classification scenarios.

All three of our proposed fusion approaches present strong capabilities in leveraging pretrained diffusion features. A more detailed comparison and interpretation analysis can be found in~\cref{sec:analysis}.

\subsection{Ablation Studies}
\label{sec:ablations}
We ablate key design choices to validate our approach. We first investigate the impact of individual diffusion features extracted at different diffusion stages. We then compare our fusion strategies against “raw” features, and also evaluate different pretrained backbones and data-scarcity scenarios.

\begin{figure}[tp]
    \centering
    \includegraphics[width=\linewidth]{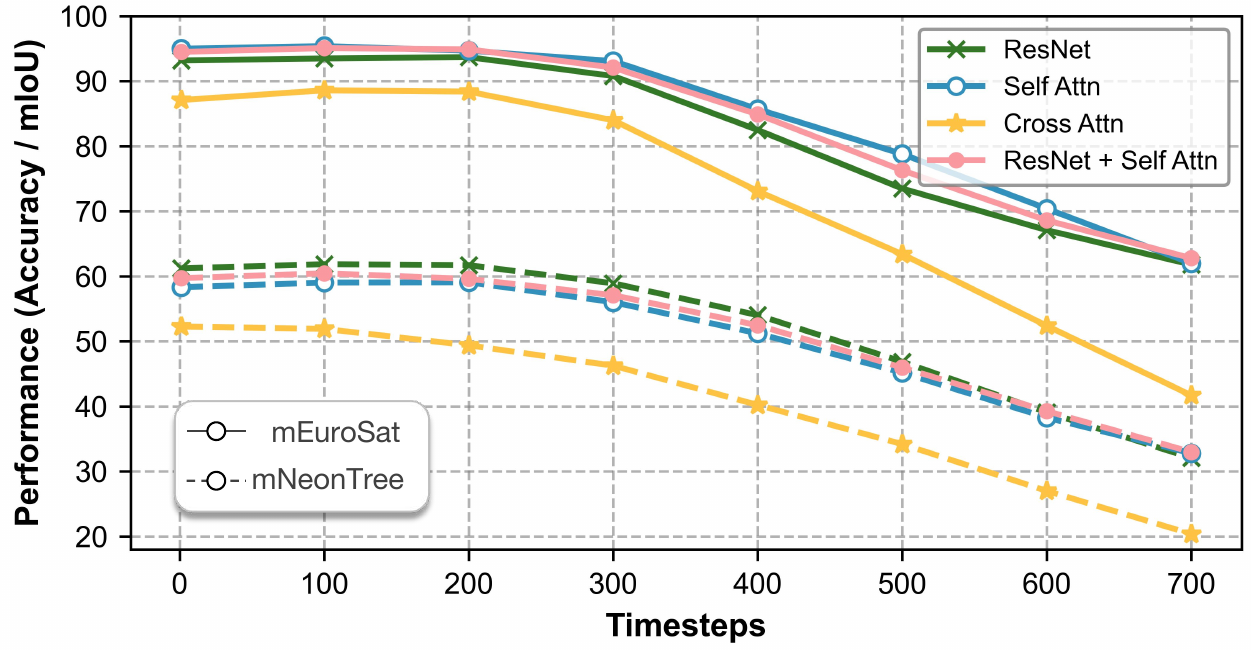}
    \vspace{-10pt}
\caption{Performance of individual module blocks across sampling timesteps on mEuroSat and mNeonTree datasets.}
\vspace{-3pt}
\label{fig:timesteps}
\end{figure}

\vspace{-3pt}
\paragraph{Effects of Timesteps.}Using one classification and one segmentation task as examples (\cref{fig:timesteps}), we observe that performance peaks when sampling within the first ~20\% of the diffusion timesteps. Typically, DDIM~\cite{song2021denoising} sampling can take up to 1000 steps. At later steps, heavily noised latents lose too many fine-grained details, while very early steps (e.g., one-step denoising) provide limited learning signals.

\vspace{-3pt}
\paragraph{Effects of Module Blocks.} As depicted in~\cref{fig:timesteps}, ResNet and self-attention outputs contribute most significantly to performance, while cross-attention blocks provide minimal benefits. This aligns with our focus on vision foundation models, as cross-attention primarily encodes task-irrelevant textual information. Based on these findings, we primarily utilize ResNet and self-attention features from the initial 20\% of timesteps.

\begin{table}[t]
    \centering\renewcommand{\arraystretch}{1.1}
    \setlength{\tabcolsep}{4pt}
    \resizebox{1.0\columnwidth}{!}{
    \begin{tabular}{ccccccc}
        \toprule
        \multirow{2}{*}{Multi-T} & \multirow{2}{*}{Multi-L} & \multirow{2}{*}{Method} & \multicolumn{2}{c}{Classification}
        & \multicolumn{2}{c}{Segmentation} 
        \\
        \cmidrule(lr){4-5} \cmidrule(lr){6-7}
        & &  & s2s  & es & cashew & pv-s \\
        \midrule
        \ding{56} & \ding{56} & ts=1, SA      & 53.6   &  94.3 & 55.3 & 92.5 \\
        \ding{56} & \ding{56} & ts=1, R       &  50.8  &  92.1 & 56.4 & 92.1\\
        \ding{56} & \ding{56} & ts=100, SA    & 52.1  &  \underline{94.7} & 54.1 & \underline{93.2} \\
        \ding{56} & \ding{56} & ts=100, R     &  50.5  &  92.4 & 57.9 & 92.6 \\
        \ding{52} & \ding{52} & simple concat &  \underline{55.4} &  94.5 & \underline{59.1} & 92.9 \\
        
        \midrule
        
        \ding{52} & \ding{52} & Global fusion  & \textbf{59.3} &  \textbf{97.7}    & \textbf{66.5} &\textbf{95.1}  \\
        \ding{52} & \ding{52} & Localized  fusion& \textbf{58.9}  &  \textbf{96.8}    & \textbf{64.8} &\textbf{95.0}  \\
        \ding{52} & \ding{52} & MoE fusion     & \textbf{58.8} &  \textbf{97.3}    & \textbf{66.1} &\textbf{95.3}  \\
        \bottomrule
    \end{tabular}
    }
    \caption{Comparison of using raw features and different feature fusion strategies. The reported numbers are top-1 accuracy or mIoU. 
    }
\label{tab:raw-compare}
\end{table}

\vspace{-3pt}
\paragraph{Raw Features vs. \ Feature Fusion.}
A straightforward way to leverage diffusion-based representations is to feed raw features (without learnable fusion networks) directly into a task-specific decoder. As shown in~\cref{tab:raw-compare}, this naive approach already matches or surpasses the performance of other pretrained RS models, reflecting the discriminative capacity of generative diffusion. However, the optimal combination of timesteps and module blocks varies across datasets—one dataset might favor features from an early timestep’s self-attention, whereas another benefits from a later timestep’s ResNet activations. Simple concatenation of features occasionally improves results but suffers from high dimensionality and inconsistent gains. By contrast, our fusion strategies consistently outperform both raw features and simple concatenation, demonstrating that principled aggregation better exploits the diverse and complementary representations learned during diffusion and is crucial for maximizing downstream performance.

\begin{table}[t]
\centering
\renewcommand{\arraystretch}{1.0}
\setlength{\tabcolsep}{6pt}
\resizebox{0.8\columnwidth}{!}{
\begin{tabular}{lcccccc}
        \toprule
      & \multicolumn{2}{c}{$E=4$}
        & \multicolumn{2}{c}{$E=8$}
        & \multicolumn{2}{c}{$E=12$} \\
    \cmidrule(lr){2-3} \cmidrule(lr){4-5} \cmidrule(lr){6-7}
     top-$k$ & 1& 2 & 1 &2 & 1 & 2 \\
    \midrule
    mIoU  &60.5	&69.8	&60.2	&\textbf{71.6}	&59.6	&68.3 \\ 
    \bottomrule
\end{tabular}
}
\caption{Ablation studies on number of experts $E$ and top-$k$ in MoE fusion on m-chesapeak-landcover dataset.}
\label{tab:moe-ablation}
\end{table}

\vspace{-3pt}
\paragraph{Number of Experts in MoE Fusion.}
We quantify the impact on the number of experts $E$ and routing parameter top-$k$ in MoE fusion method. \cref{tab:moe-ablation} shows that fewer experts lead to insufficient learning, while more experts may introduce redundancy. Using top-$k$=2 consistently outperforms a single expert activation, suggesting the benefit of complementary expert representations. We select $E=8$ and top-$k$=2 as our configuration to balance computational efficiency and performance.

\begin{table}[t]
\centering
\renewcommand{\arraystretch}{1.05}
\setlength{\tabcolsep}{7pt}
\resizebox{1.0\columnwidth}{!}{
\begin{tabular}{lcccccc}
    \toprule
    \multirow{2}{*}{Backbone} & \multicolumn{3}{c}{Binary-class}
        & \multicolumn{3}{c}{Multi-class} \\
    \cmidrule(lr){2-4} \cmidrule(lr){5-7} 
     & pv-s & nz-c & neon & cashew & sa-c & ches \\
    \midrule
    SD v2-1         & 94.5 & 82.5 &60.2 & 63.7  &29.3  & 65.8\\
    DiffusionSat    & \textbf{95.1}	&\textbf{83.5}  &\textbf{61.8} & \textbf{66.5}	&\textbf{32.6}  & \textbf{69.5} \\
    \bottomrule
\end{tabular}
}
\vspace{-5pt}
\caption{Performance comparison of diffusion backbones using global weighted fusion strategy on segmentation tasks.}

\label{tab-sd21}
\end{table}

\vspace{-3pt}
\paragraph{Comparison of Diffusion Backbones.} 

In~\cref{tab-sd21}, we compare two pretrained diffusion backbones for semantic segmentation tasks using the global weighted fusion strategy. \emph{SD v2-1}~\cite{Rombach_2022_CVPR} is trained on a large-scale web-scraped dataset, while \emph{DiffusionSat}~\cite{khanna2024diffusionsat} further finetunes it using large-scale satellite imagery. Despite not being domain-specific, SD v2-1 still achieves competitive results, likely due to its massive and diverse pretraining set. However, the specialized DiffusionSat model consistently outperforms SD v2-1, particularly on more complex multi-class tasks, highlighting the benefits of domain-focused finetuning.

\paragraph{Data Scarcity Scenario.}
\begin{table}[tb!]
\centering
\renewcommand{\arraystretch}{1.05}
\setlength{\tabcolsep}{5pt}
\resizebox{1.0\columnwidth}{!}{
\begin{tabular}{lcccccc}
        \toprule
    \multirow{2}{*}{Method} & \multicolumn{2}{c}{pv-s}
        & \multicolumn{2}{c}{nz-c}
        & \multicolumn{2}{c}{cashew} \\
    \cmidrule(lr){2-3} \cmidrule(lr){4-5} \cmidrule(lr){6-7}
     & 100\% & 10\% & 100\% & 10\% & 100\% & 10\% \\
    \midrule
    Satlas~\cite{bastani2023satlaspretrain} &92.3	&88.6	&83.1	&77.8	&49.1	&25.7  \\
    SSL4EO-MAE~\cite{ssl4eo}               &89.2	&84.8	&78.7	&68.1	&57.8	&26.8 \\
    ScaleMAE~\cite{reed2023scalemae}      & 94.2	&91.3	&\underline{\colorbox{colorFst}{\bf 84.1}}	&78.4	&47.8	&27.8 \\
    SSL4EO-DINO~\cite{ssl4eo}             &89.0	&85.7	&78.9	&66.2	&61.3	&31.2  \\
    GFM~\cite{mendieta2023towards}         & 93.1	 &90.3	&82.4	 &74.1	&53.5	 &22.8 \\
    RemoteCLIP~\cite{mall2024remote}       &93.2	&90.0	&80.7	&73.7	&51.7	&27.1\\
    CROMA~\cite{fuller2023croma}           &92.5	&88.7	&83.4	&75.4	&\underline{62.2}	&\underline{34.3} \\
    DOFA~\cite{xiong2024neural}            & \underline{94.8}	&\underline{92.3}	&82.8	&\underline{79.4}	&53.9	&29.5\\
    \midrule
    \multicolumn{7}{l}{\cellcolor[HTML]{EEEEEE}{\textit{\ourmodel (Ours)}}} \\
    Global fusion   &\colorbox{colorSnd}{95.1}	&\colorbox{colorSnd}{93.5}	&\colorbox{colorTrd}{83.5}	&\colorbox{colorSnd}{80.0}	&\colorbox{colorFst}{\bf66.5}	&\colorbox{colorFst}{\bf39.6}\\
    Localized  fusion & \colorbox{colorTrd}{95.0}	& \colorbox{colorTrd}{93.3}	&83.2	&\colorbox{colorTrd}{79.5}	&\colorbox{colorTrd}{64.8}	&\colorbox{colorTrd}{38.1} \\
    MoE fusion      &\colorbox{colorFst}{\bf 95.3}	&\colorbox{colorFst}{\bf 93.9}	&\colorbox{colorSnd}{83.7}	&\colorbox{colorFst}{\bf80.3}	&\colorbox{colorSnd}{66.1}	&\colorbox{colorSnd}{38.5} \\
    \bottomrule
\end{tabular}
\vspace{-2pt}}
\caption{Semantic segmentation performance when using 100\% and 10\% of labeled data on the m-pv4ger-seg, m-nz-cattle and m-cashew-plantation datasets.}
\label{tab:limited_ss1}
\end{table}

We further evaluate our method under data scarcity by reducing the labeled training data to 10\% while preserving the original data distribution. Results in~\cref{tab:limited_ss1} show that {\ourmodel} maintains robust performance in limited-data scenarios, demonstrating its generalization capability even with fewer labeled samples.

\subsection{Qualitative Analysis}
\label{sec:analysis}

\noindent\textbf{Feature Visualization.}
\begin{figure}[tp]
    \centering
    \includegraphics[width=\linewidth]{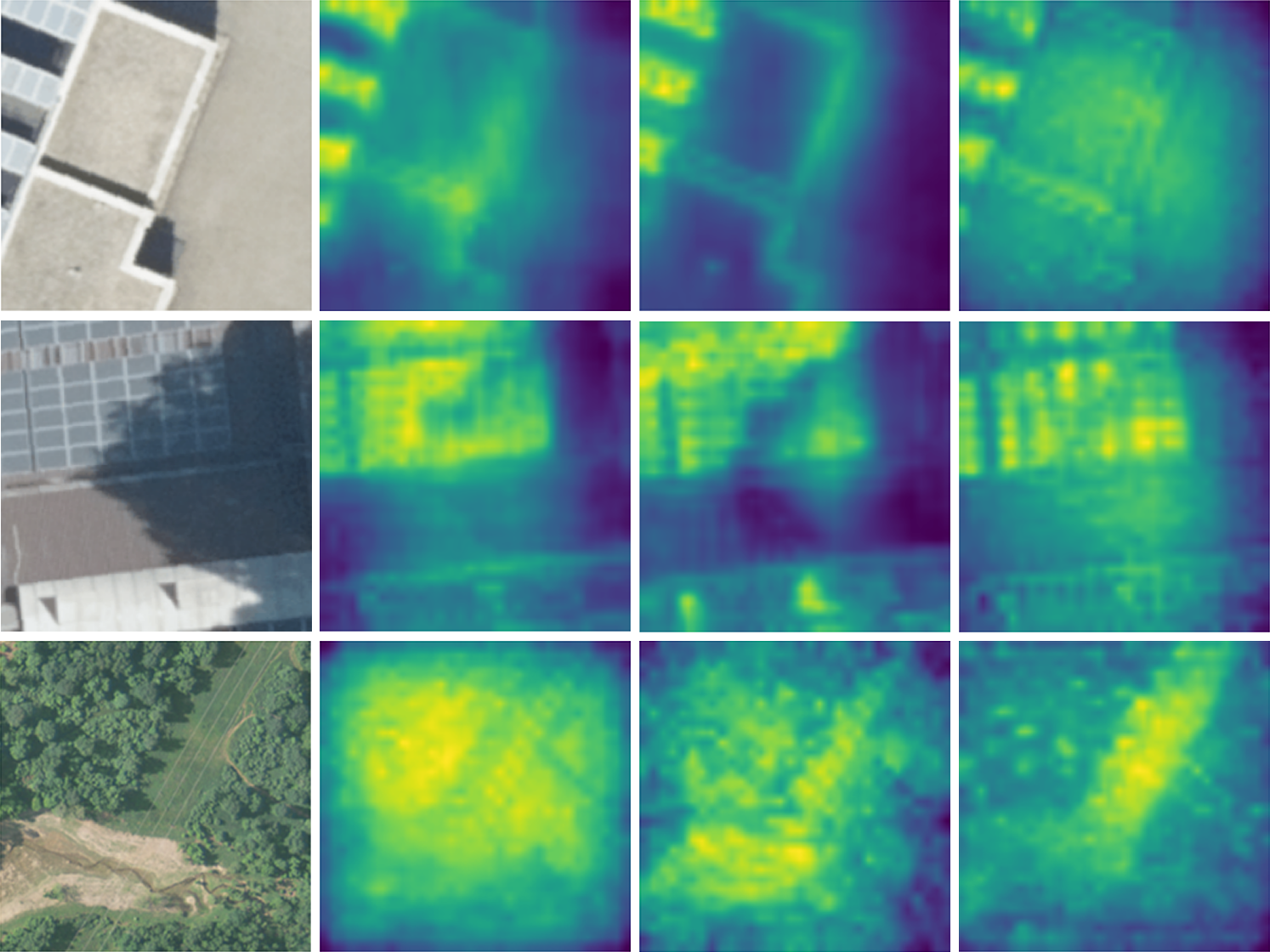}
    {\footnotesize
    \begin{tabularx}{\linewidth}{*{4}{Y}}
    Image & (a) G-W & (b) L-W & (c) MoE \\
    \end{tabularx}
    }
    \vspace{-10pt}
    \caption{Visualization of \textbf{fused feature maps} obtained by the three fusion strategies, demonstrating their distinct emphasis. G-W and L-W denote global weighted and localized weighted fusion.}
    \vspace{-1pt}
    \label{fig:fused-feats}
\end{figure}

\begin{figure}[tp]
    \centering
    \includegraphics[width=\linewidth]{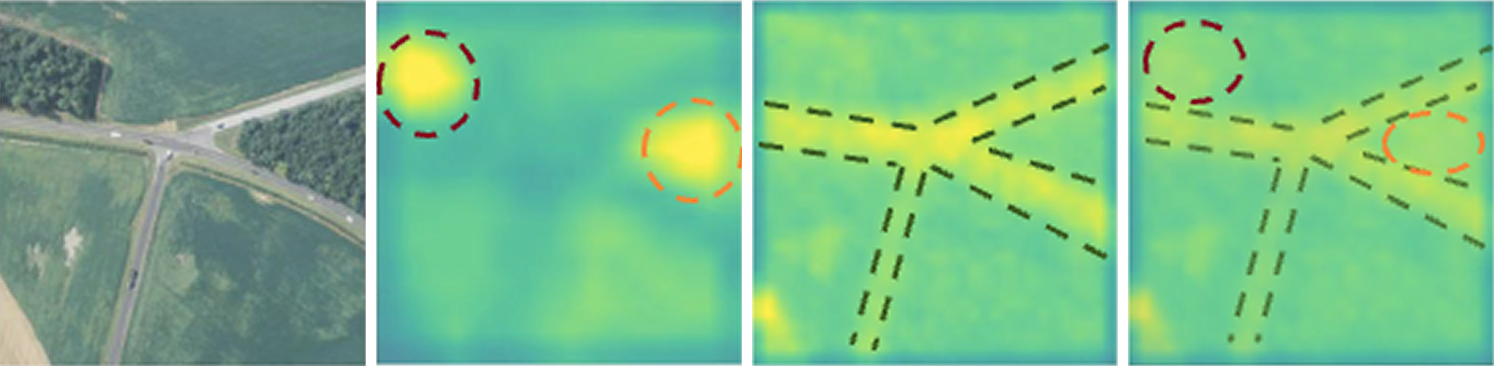}
    {\small
    \begin{tabularx}{\linewidth}{*{4}{Y}}
    (a) Image & (b) Expert-1 & (c) Expert-2 & (d) Fused \\
    \end{tabularx}
    }
    \vspace{-10pt}
\caption{Visualization of individual \textbf{expert outputs} in the MoE fusion strategy, showing each expert specializing in distinct spatial patterns and textures.}
\vspace{-10pt}
\label{fig:moe-experts}
\end{figure}

\cref{fig:fused-feats} visualizes the fused feature maps learned by {\ourmodel}'s three fusion methods, highlighting their distinct characteristics. 
Localized weighted fusion produces detailed maps that can emphasize fine-grained object boundaries  (e.g., buildings in the first row (b)), but it is also sensitive to local variations, such as shadows commonly observed in RS images (second row (b), split appearance of buildings). In contrast, Global Weighted fusion yields more stable representations across the shadowed regions, yet it may present limitations in preserving certain fine-grained details (as shown in the third row(a)).
The MoE joint fusion balances these trade-offs by dynamically activating specialized expert sub-networks, preserving both global contexts and local details while mitigating sensitivity to perturbations (see the second row(c)). As further illustrated in~\cref{fig:moe-experts}, each expert learns distinct spatial or textural patterns, highlighting their abilities to adapt to the complexity of remote sensing imagery.

\noindent\textbf{Prediction Results.}
\begin{figure}[tp]
    \centering
    \includegraphics[width=\linewidth]{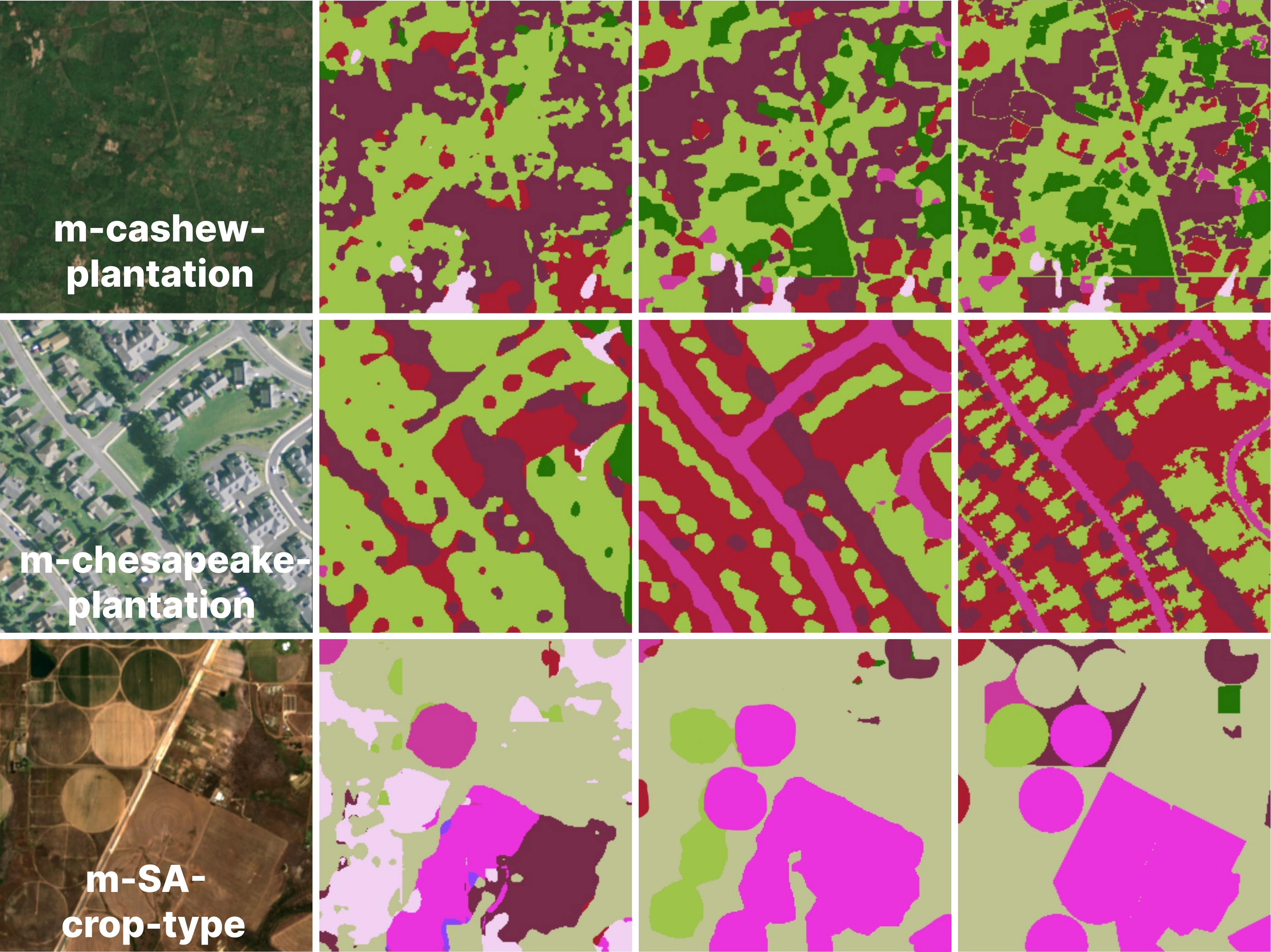}
    
    {\small \begin{tabularx}{\linewidth}{*{4}{Y}}
        (a) Image & (b) Others & (c) Ours & (d) Label \\
    \end{tabularx} }
    \vspace{-10pt}
    \caption{\textbf{Example segmentation predictions} comparing {\ourmodel} (ours) and the best-performing pretrained RS baselines per dataset. For m-cashew-plantation dataset, we colorize the clasees as follows: \textcolor[RGB]{34, 114, 4}{well-managed plantation}, \textcolor[RGB]{118, 44, 73}{poorly-managed plantation}, \textcolor[RGB]{168, 29, 47}{non-plantation}, \textcolor[RGB]{241, 210, 242}{residential}, \textcolor[RGB]{158, 196, 74}{background}, and \textcolor[RGB]{204, 57, 156}{uncertain}.
For m-chesapeake-landcover dataset, the color maps are: \textcolor[RGB]{118,44,73}{tree-canopy-forest}, \textcolor[RGB]{168,29,47}{low-vegetation-field}, \textcolor[RGB]{241,210,242}{barren land}, \textcolor[RGB]{158,196,74}{impervious-other}, \textcolor[RGB]{204, 57, 166}{impervious-roads}. For m-SA dataset, the classes are represented as: \textcolor[RGB]{34, 114, 4}{lucerne/medics}, \textcolor[RGB]{118, 44, 73}{planted pastures}, \textcolor[RGB]{168, 29, 47}{fallow}, \textcolor[RGB]{241, 210, 242}{wine grapes}, \textcolor[RGB]{158, 196, 74}{weeds}, \textcolor[RGB]{204, 57, 156}{canola} and \textcolor[RGB]{235, 51, 221}{rooibos}.}
    \label{fig:pred-examples}
    \vspace{-10pt}
\end{figure}

\cref{fig:pred-examples} presents examples of semantic segmentation results obtained from our top-performing model and compared to the best-performing alternative GFMs on m-cashew-plantation, m-chesapeak-landcover and m-SA-crop-type. Our model demonstrates strong segmentation capability in distinguishing between different land cover types, showcasing its effectiveness in transferring robust representation for various RS downstream tasks.

\section{Conclusion}
\label{sec:conclusion}
we introduced {\ourmodel}, a novel framework that systematically adapts diffusion-based generative models for self-supervised representation learning in remote sensing. By leveraging multi-stage diffusion features and novel fusion strategies, {\ourmodel} achieves state-of-the-art performance on various discriminative RS tasks, demonstrating diffusion models’ viability as scalable SSL alternatives.

Despite the great advantages, some limitations remain. First, our study focuses on RGB imagery due to limited pretraining on multi-band data of the utilized backbone.
Extending {\ourmodel} to multi-modal RS data 
could unlock further capabilities. 
Additionally, while we primarily leverage unconditional diffusion models,
investigating conditioned image-to-image diffusion models—given their dense guidance mechanisms—could further enhance representations. Addressing these aspects in future work will contribute to a more comprehensive integration of diffusion models into GFMs framework, further bridging the gap between generative and discriminative paradigms in RS.

{\small
\section*{Acknowledgements}
This project is funded by the Division of Geoinformatics at KTH Royal Institute of Technology, the Geomatics section at KU Leuven and the Natural Science Foundation of Shanghai (Grant No. 25ZR1402268). We also thank the National Academic Infrastructure for Supercomputing in Sweden (NAISS, Grant No. 2022-06725) for supporting the computations and data handling.
} 

{
    \small
    \bibliographystyle{ieeenat_fullname}
    \bibliography{main}
}

\clearpage
\setcounter{page}{1}
\maketitlesupplementary

\section{Diffusion Models}
\label{supp:diff-equations}
Below, we provide a concise mathematical overview of discrete diffusion models (DMs).

\subsection{Diffusion Model Essentials}

Let $\mathbf{x}_0 \sim p(\mathbf{x}_0)$ be a sample from an underlying data distribution. A forward diffusion process iteratively adds Gaussian noise over $T$ discrete timesteps, producing corrupted samples $\mathbf{x}_1, \ldots, \mathbf{x}_T$. One common choice to model each step is:
\begin{equation}
p(\mathbf{x}_t \mid \mathbf{x}_{t-1})
= \mathcal{N}\!\bigl(\mathbf{x}_t;\,\sqrt{1-\beta_t}\,\mathbf{x}_{t-1},\,\beta_t\,\mathbf{I}\bigr),
\end{equation}
where $\beta_t\in(0,1)$ controls the noise variance. One can obtain $\mathbf{x}_t$ directly from the original image $\mathbf{x}_0$ given the cumulative effect of t nose-adding steps:
\begin{equation} \label{eq:add-noise}
\mathbf{x}_t
= \sqrt{\bar{\alpha}_t}\,\mathbf{x}_0 + \sqrt{1-\bar{\alpha}_t}\,\boldsymbol{\epsilon},
\quad
\bar{\alpha}_t = \prod_{i=1}^t{\alpha_i},  
\end{equation}
where $\alpha_t=1-\beta_t$ and $\boldsymbol{\epsilon}\sim \mathcal{N}(\mathbf{0},\mathbf{I})$. As $t$ grows, $\mathbf{x}_t$ becomes increasingly noisy; at $t=T$, the corrupted $\mathbf{x}_T$ approximates a pure Gaussian distribution, losing most structure of $\mathbf{x}_0$.

To generate novel samples starting from pure noise, a diffusion model learns a reverse denoising process $p_\theta(\mathbf{x}_{t-1} \mid \mathbf{x}_t)$, parameterized by $\theta$, which conceptually “denoises” $\mathbf{x}_t$ step by step until recovering $\mathbf{x}_0$:
\begin{equation}
p_\theta(\mathbf{x}_{t-1} \mid \mathbf{x}_t)
= \mathcal{N}\!\bigl(\mathbf{x}_{t-1};\,\boldsymbol{\mu}_\theta(\mathbf{x}_t,t),\,\Sigma_\theta(\mathbf{x}_t,t)\bigr).
\end{equation}

A common training objective is to minimize the distance between the true noise $\boldsymbol{\epsilon}_t$ and the model’s predicted noise $\boldsymbol{\epsilon}_\theta(\mathbf{x}_t,t)$:
\begin{equation}
\mathcal{L}_\text{simple}(\theta)
\mathbb{E}_{\mathbf{x}_0,\boldsymbol{\epsilon},t}
\bigl[
\|\boldsymbol{\epsilon}_t - \boldsymbol{\epsilon}_\theta(\mathbf{x}_t,t)\|^2
\bigr].
\end{equation}

At inference, sampling proceeds from $\mathbf{x}_T \sim \mathcal{N}(\mathbf{0},\mathbf{I})$ and iteratively applies $p_\theta$ to yield a final $\mathbf{x}_0$.

\subsection{Feature Extraction using Diffusion Models}
While diffusion models (DMs) are primarily designed for image generation from Gaussian noise, our goal is to extract their learned representations for real images. To achieve this, we first \emph{invert} a real image into a noisy state and then perform the reverse denoising process.

To illustrate the inversion step, we revisit DDIM~\cite{song2021denoising}, a widely adopted sampling approach known for faster generation and invertibility. A common deterministic formulation for going from $\mathbf{x}_t$ to $\mathbf{x}_{t-1}$ is:
\begin{equation}
\mathbf{x}_{t-1} = \sqrt{\alpha_{t-1}}\hat{\mathbf{x}_0}
	+\sqrt{1 - \alpha_{t-1}}\;\epsilon_\theta(\mathbf{x}_t,t).
\end{equation}
where 
$\hat{\mathbf{x}_0}=\frac{\mathbf{x}_t - \sqrt{1-\alpha_t}\,\epsilon_\theta(\mathbf{x}_t, t)}{\sqrt{\alpha_t}}$ 
is the predicted clean image. 
By removing explicit Gaussian noise additions at each step, the process becomes deterministic, allowing a “mirror pass” that encodes $\mathbf{x}_0$ to $\mathbf{x}_T$. If we then use $\mathbf{x}_T$ as the start of the usual sampling procedure, we recover the original $\mathbf{x}_0$.

We leverage this property by reversing the order of timesteps, going from $\mathbf{x}_{t-1}$ to $\mathbf{x}_t$, starting from $\mathbf{x}_0$, and then running the denoising process on $\mathbf{x}_t$ to extract features.

While one could alternatively introduce noise into a real image by selecting a timestep and manually adding noise via~\cref{eq:add-noise}, this approach introduces stochastic variations. To ensure consistency, we adopt DDIM inversion, leveraging its deterministic nature to repurpose diffusion models for discriminative tasks.

\section{Datasets Description}
\label{supp:datsets-details}
There are six classification datasets in GEO-Bench~\cite{lacoste2024geo}:

\noindent\textbf{m-bigearthnet}
It contains 120 $\times$ 120 images with 43 land cover classes. The dataset includes 20,000 training samples, 1,000 validation samples, and 1,000 test samples. It consists of 12 spectral bands obtained from Sentinel-2 imagery, with a spatial resolution of 10.0m for the RGB channels.

\noindent\textbf{m-brick-kiln}
It consists of 64 $\times$ 64 images with 2 classes, focusing on brick kiln detection. The dataset includes 15,063 training samples, 999 validation samples, and 999 test samples. The imagery is derived from Sentinel-2 with 10 spectral bands and a resolution of 10.0m for RGB. Additional Sentinel-1 data is included. 

\noindent\textbf{m-eurosat}
It consists of 64 $\times$ 64 images spanning 10 land cover classes. The dataset contains 2,000 training samples, 1,000 validation samples, and 1,000 test samples. It includes 13 spectral bands captured from Sentinel-2 with an RGB resolution of 10.0m. 

\noindent\textbf{m-forestnet}
This dataset contains 332 $\times$ 332 images and covers 12 classes related to forest monitoring. It includes 6,464 training samples, 989 validation samples, and 993 test samples. The dataset comprises 6 spectral bands obtained from Landsat-8, with a spatial resolution of 15.0m for the RGB channels.

\noindent\textbf{m-pv4ger}
This dataset comprises 320 $\times$ 320 images covering 2 classes, with 11,814 training samples, 999 validation samples, and 999 test samples. The imagery is obtained from RGB data, with a spatial resolution of 0.1m.

\noindent\textbf{m-so2sat}
This dataset consists of 32 $\times$ 32 images spanning 17 different land cover classes. It contains 19,992 training samples, 986 validation samples, and 986 test samples. The images are derived from Sentinel-2 data with 13 spectral bands and a spatial resolution of 10.0m for RGB.

In addition, six semantic segmentation datasets are included:

\noindent\textbf{m-pv4ger-seg}
It is the segmentation variant of m-pv4ger, containing 320 $\times$ 320 images with 3,000 training samples, 403 validation samples, and 403 test samples. The dataset has 3 spectral bands (RGB) with a spatial resolution of 0.1m.

\noindent\textbf{m-nz-cattle}
It contains 500 $\times$ 500 images with 2 classes, including 524 training samples, 66 validation samples, and 65 test samples. The imagery consists of 3 spectral bands (RGB) with an unknown spatial resolution.

\noindent\textbf{m-NeonTree}
It includes 400 $\times$ 400 images with 2 classes, consisting of 270 training samples, 94 validation samples, and 93 test samples. The dataset comprises 5 spectral bands (RGB + Hyperspectral + Elevation).

 \noindent \noindent\textbf{m-cashew-plantation}
It comprises 256 $\times$ 256 images with 7 classes, featuring 1,350 training samples, 400 validation samples, and 400 test samples. The imagery is sourced from Sentinel-2 with 10 spectral bands and an RGB resolution of 10.0m. 

\noindent\textbf{m-SA-crop-type}
This dataset consists of 256 $\times$ 256 images with 10 classes. It contains 3,000 training samples, 1,000 validation samples, and 1,000 test samples. The imagery is sourced from Sentinel-2 with 10 spectral bands and an RGB resolution of 10.0m.

\noindent\textbf{m-chesapeake-landcover}
This dataset consists of 256 $\times$ 256 images with 7 land cover classes. It contains 3,000 training samples, 1,000 validation samples, and 1,000 test samples. The dataset includes 4 spectral bands (RGBN) with a spatial resolution of 1.0m.

\section{Evaluation Details}
\label{supp:eval-details}
We provide additional details regarding the evaluation process in this section

For task training criteria, we utilize nn.CrossEntropyLoss() from the PyTorch library for all tasks, except for the m-bigearthnet dataset, which follows a multi-label classification setup and nn.BCEWithLogitsLoss() is applied.

Regarding training schedules, all models are trained for 60 epochs on the m-cashew-plantation and m-sa-crop-type datasets, while the remaining datasets undergo training for 40 epochs.

Additionally, we apply data augmentation techniques for all datasets to enhance model generalization. During training, images undergo random horizontal flipping, vertical flipping, and color jittering based on a probabilistic threshold of 0.5.

\section{Visualizations of Global Weighting}
\label{sec:global-weighting-supp}
\begin{figure}[bp]
    \centering
    \begin{subfigure}{\columnwidth}
        \centering
        \includegraphics[width=\columnwidth]{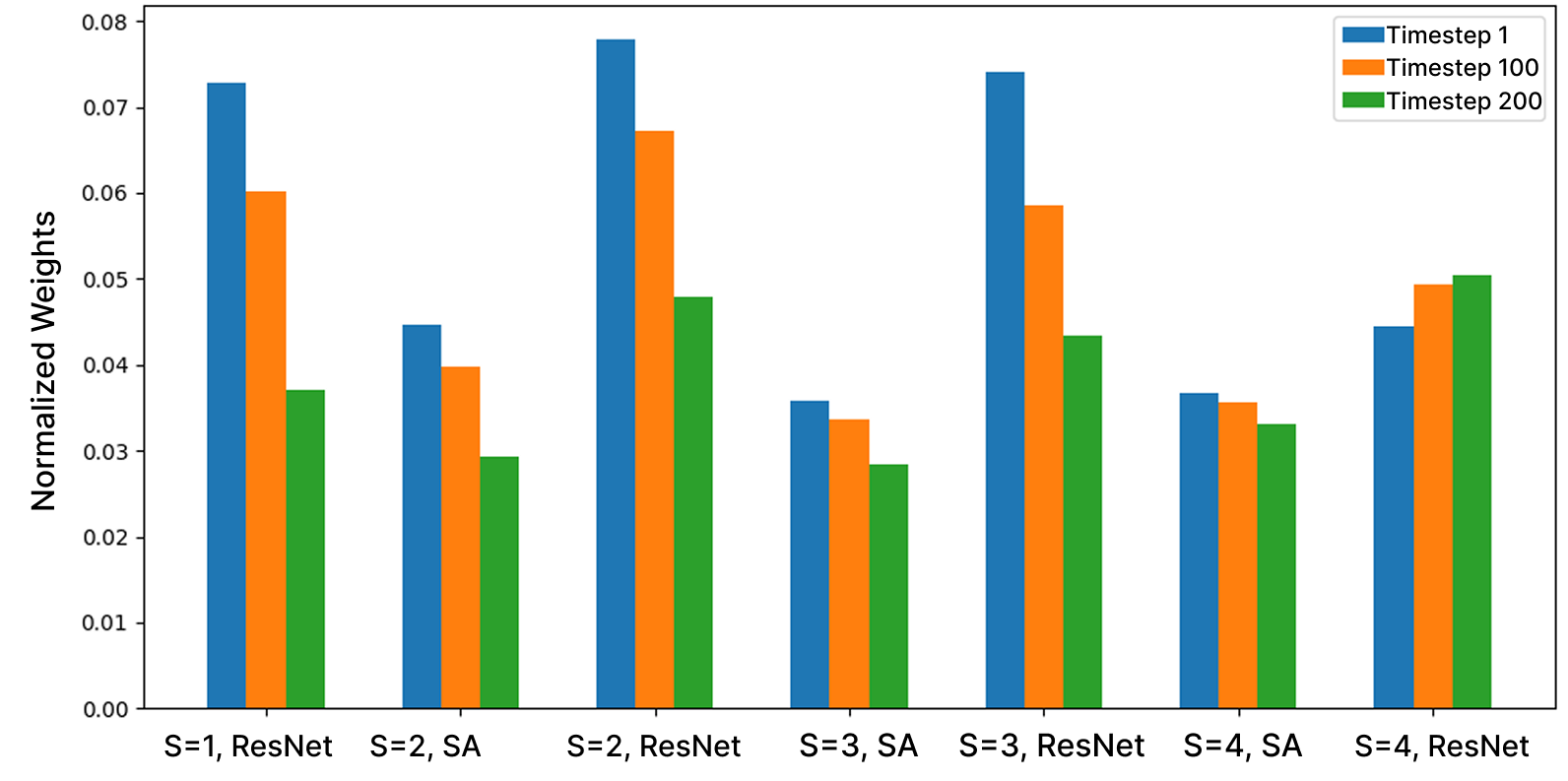}
        \caption{Weight allocations for m-chesapeake-landcover dataset.}
    \end{subfigure}
    
    \vspace{5pt} %

    \begin{subfigure}{\columnwidth}
        \centering
        \includegraphics[width=\columnwidth]{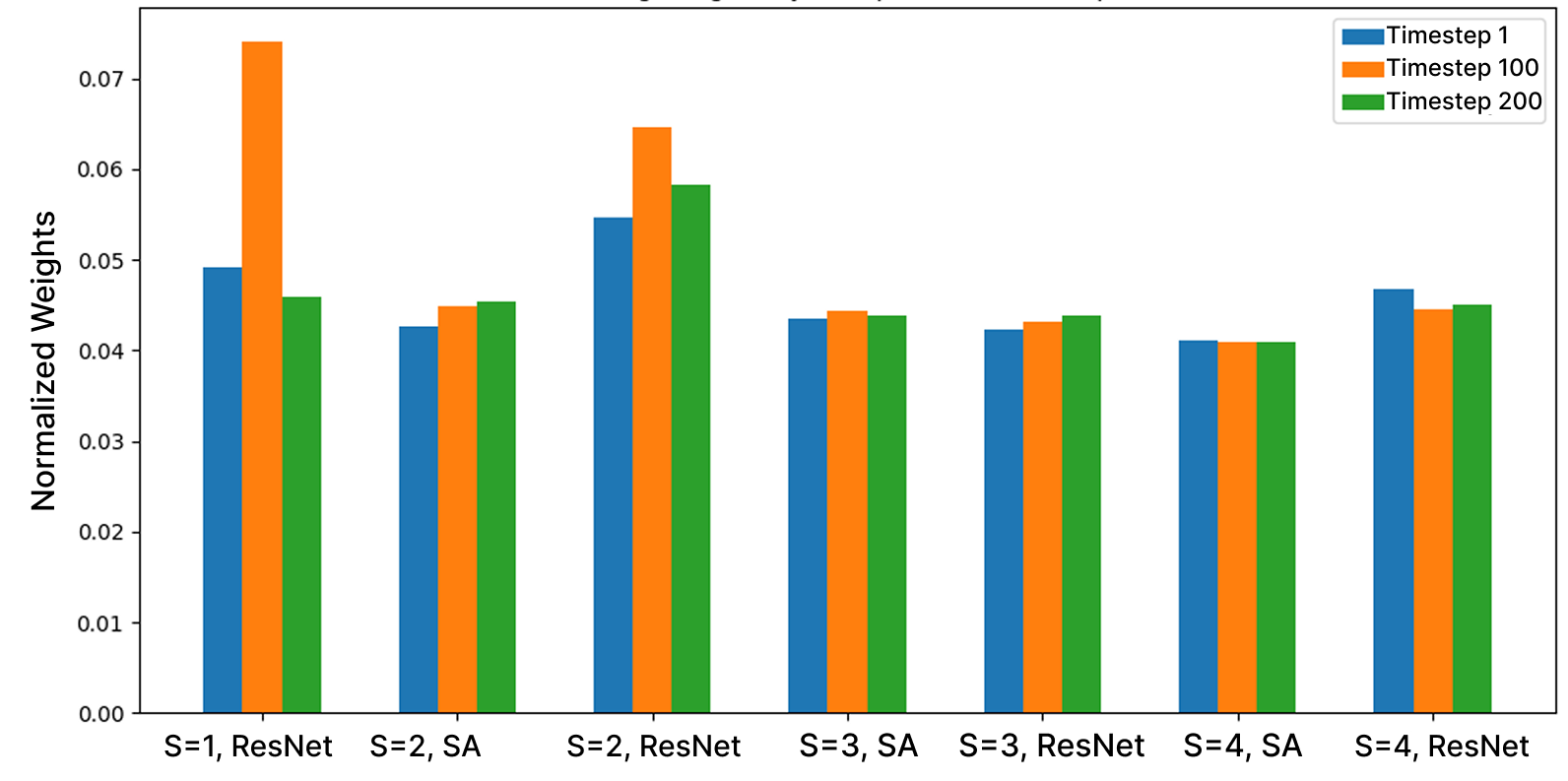}
        \caption{Weight allocations for m-pv4ger-seg dataset.}
    \end{subfigure}

    \begin{subfigure}{\columnwidth}
        \centering
        \includegraphics[width=\columnwidth]{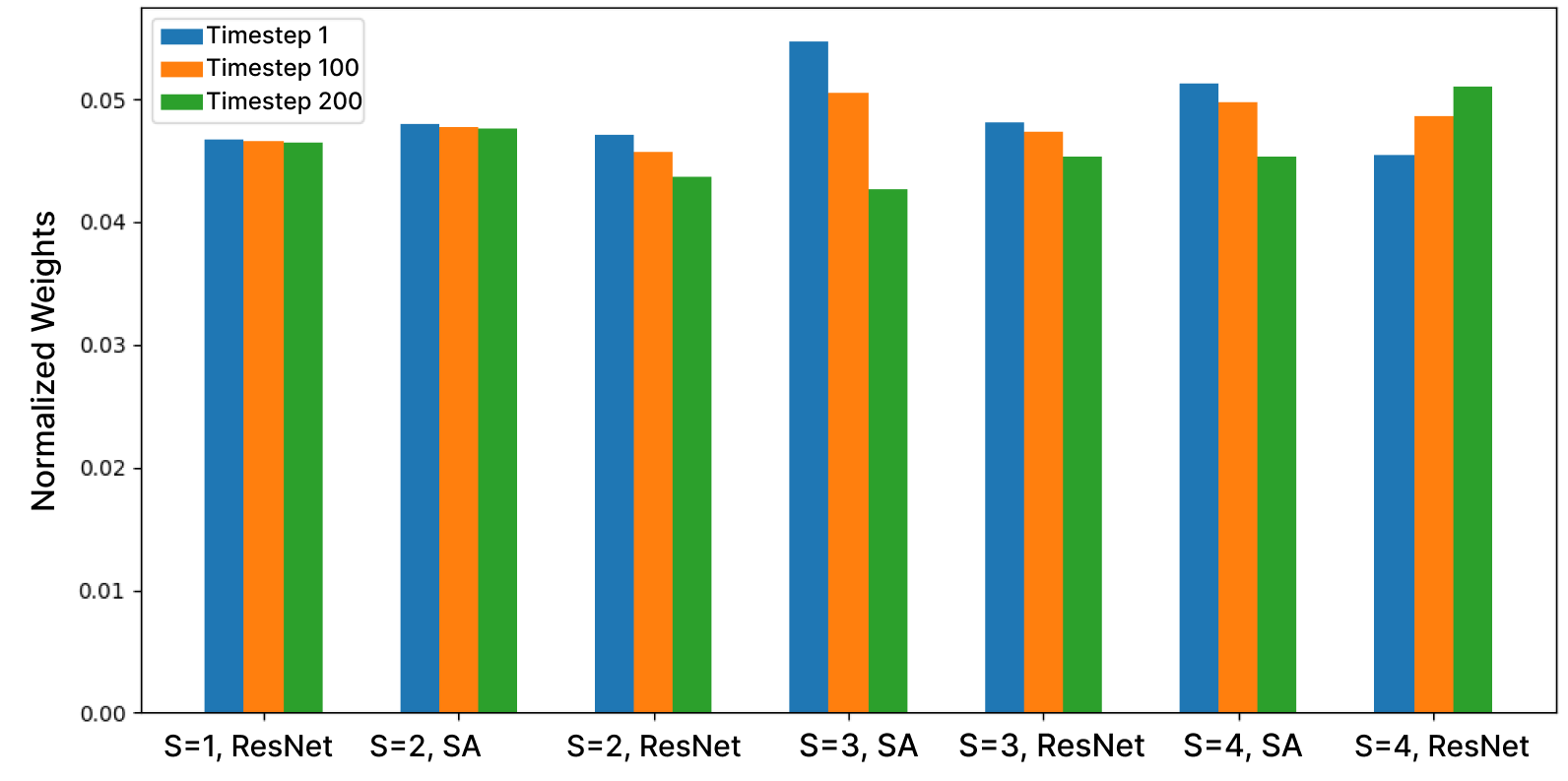}
        \caption{Weight allocations for m-NeonTree dataset.}
    \end{subfigure}
    
    \caption{Normalized weight allocations in Global Weighted Fusion across different datasets.}
    \label{fig:global-weights-dist}
\end{figure}

In~\cref{sec:ablations}, we demonstrate that features from different blocks and timesteps contribute differently depending on the dataset. Our global weighted fusion method effectively aggregates these features to enhance performance. \cref{fig:global-weights-dist} visualizes the learned weight distributions across blocks and timesteps at different scales, illustrating how our fusion strategy dynamically adjusts feature importance for each dataset. This automated weighting helps reduce the need for manual feature selection, promoting adaptive and optimal feature integration.

\end{document}